\documentclass[10pt,twocolumn,letterpaper]{article}

\usepackage{cvpr}              

\usepackage[dvipsnames]{xcolor}

\definecolor{cvprblue}{rgb}{0.21,0.49,0.74}
\usepackage[pagebackref,breaklinks,colorlinks,citecolor=cvprblue]{hyperref}
\usepackage{pifont,algorithm,algorithmic}
\usepackage{amsmath}
\usepackage{float}
\usepackage{tabularx}
\usepackage{multirow}
\usepackage{bm}
\def\paperID{774} 
\def\confName{CVPR}
\def\confYear{2024}

\title{Consistent Prompting for Rehearsal-Free Continual Learning}

\author{Zhanxin Gao\textsuperscript{1}, Jun Cen\textsuperscript{2}, Xiaobin Chang\textsuperscript{1,3}\thanks{indicates corresponding author.}\\
$^1$School of Artificial Intelligence, Sun Yat-sen University, China\\
$^2$Cheng Kar-Shun Robotics Institute, The Hong Kong University of Science and Technology, China\\
$^3$Key Laboratory of Machine Intelligence and Advanced Computing, Ministry of Education, China\\
{\tt\small gaozhx27@mail2.sysu.edu.cn, jcenaa@connect.ust.hk, changxb3@mail.sysu.edu.cn}
}

\begin{document}
\maketitle
\begin{abstract}
Continual learning empowers models to adapt autonomously to the ever-changing environment or data streams without forgetting old knowledge. Prompt-based approaches are built on frozen pre-trained models to learn the task-specific prompts and classifiers efficiently. Existing prompt-based methods are inconsistent between training and testing, limiting their effectiveness. Two types of inconsistency are revealed. Test predictions are made from all classifiers while training only focuses on the current task classifier without holistic alignment, leading to Classifier inconsistency. Prompt inconsistency indicates that the prompt selected during testing
may not correspond to the one associated with this task during training. In this paper, we propose a novel prompt-based method, Consistent Prompting (CPrompt), for more aligned training and testing. Specifically, all existing classifiers are exposed to prompt training, resulting in classifier consistency learning. In addition, prompt consistency learning is proposed to enhance prediction robustness and boost prompt selection accuracy. Our Consistent Prompting surpasses its prompt-based counterparts and achieves state-of-the-art performance on multiple continual learning benchmarks.
Detailed analysis shows that improvements come from more consistent training and testing. Our code is available at \url{https://github.com/Zhanxin-Gao/CPrompt}.

\end{abstract}    
\section{Introduction}
\label{sec:intro}

Continual learning~\cite{sodhani2022introduction,de2021continual,masana2022class,belouadah2021comprehensive,van2019three} aims to equip deep models with the capacity to continuously acquire new knowledge, e.g., learn to recognize new object categories, while handling catastrophic forgetting~\cite{mccloskey1989catastrophic,lee2017overcoming,ramasesh2021effect,mehta2021empirical} of the old knowledge. Rehearsal-based methods~\cite{rebuffi2017icarl,bonicelli2022effectiveness,yoon2021online} alleviate the forgetting problem with a small number of exemplars of previous tasks stored in the memory buffer and replayed with the new task data during training. However, exemplars of previous tasks may not be available due to constraints such as data privacy or memory limitations. Therefore, rehearsal-free continual learning methods~\cite{ma2022progressive,smith2021always,gao2022r,liu2022few,zhang2023slca,wang2024hierarchical} have attracted much attention.

\begin{figure}[t]
  \centering
  \centerline{\includegraphics[width=0.99\linewidth]{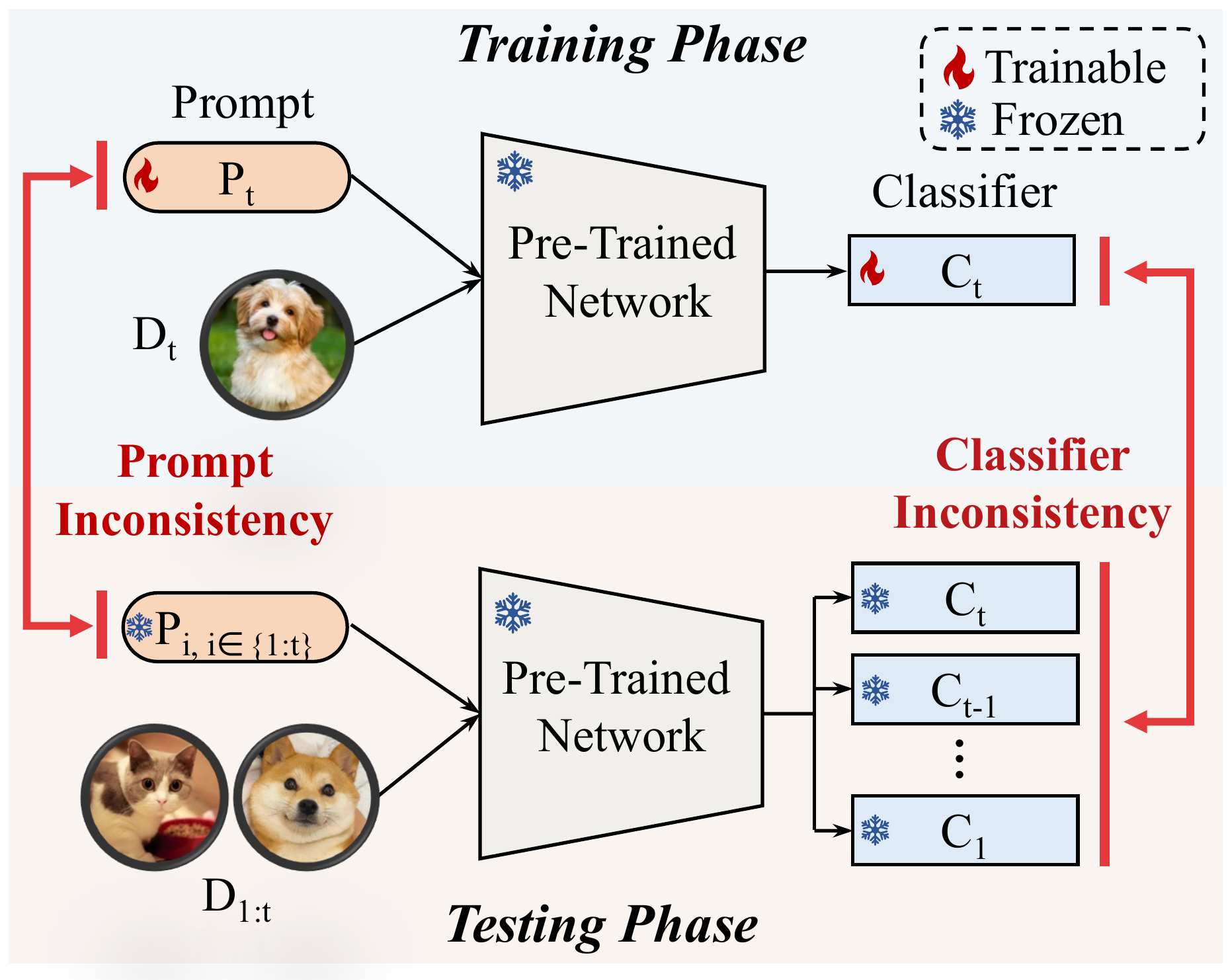}}
\caption{
Existing prompt-based methods are inconsistent between training and testing, and prompt inconsistency and classifier inconsistency are illustrated. $P_t$ and $C_t$ represent the prompt and the classifier of the current task t, respectively.
}
\label{fig:firstpic}
\end{figure}

Prompt-based models~\cite{wang2022learning,wang2022dualprompt,smith2023coda,pei2023space,tang2023prompt} have shown exceptional results in rehearsal-free continual learning.
Based on the frozen backbone network pre-trained on a large-scale dataset, such models can efficiently adapt to the new task by only training a tiny set of parameters, i.e., the prompts and fully connected (FC) classifiers. However, the training and testing of existing approaches lack consistency, as the prompt and the classifier are optimized solely within the current task.
As shown in Figure~\ref{fig:firstpic}, two types of such inconsistencies are discussed.
Firstly, classifier inconsistency indicates that test predictions are made from all classifiers rather than only from the training one.
Secondly, prompt inconsistency refers to the mismatch of prompts between training and testing.
Existing prompt-based methods do not identify or handle such inconsistent issues, leading to suboptimal performance.

\begin{figure}[t]
  \centering
  \centerline{\includegraphics[width=0.99\linewidth]{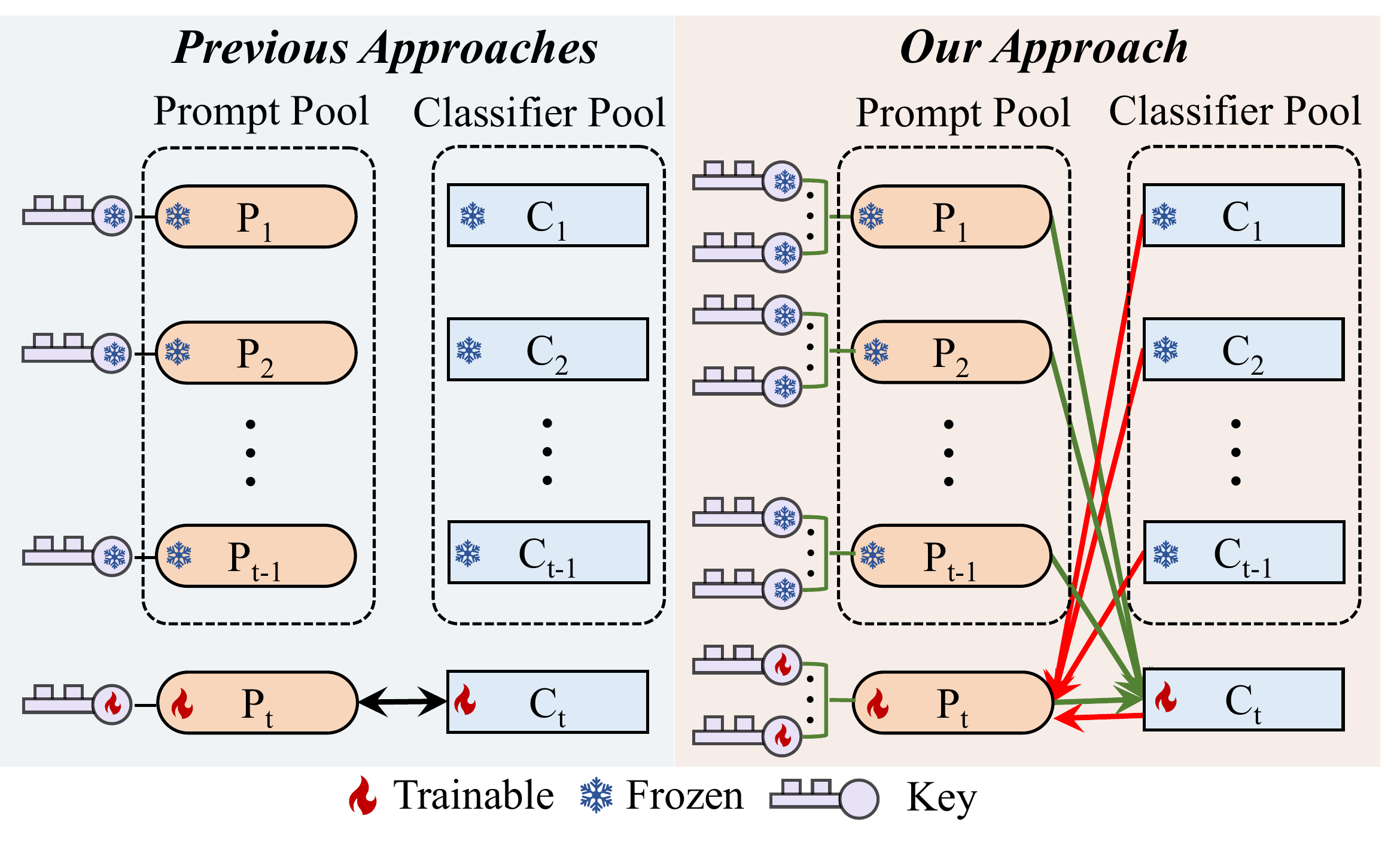}}
\caption{
Previous approaches typically train the current task prompt and classifier in isolation.
Our Consistent Prompting leverages all existing prompts and classifiers to instruct the training of the current task prompt and classifier. Meanwhile, we suggest using multiple keys to map each task prompt, instead of relying on a single key, to adapt to the diverse nature of each task.
}
\label{fig:secondpic}
\end{figure}

To align the training of prompts and classifiers with testing, we propose a novel training method, called Consistent Prompting (CPrompt), which surpasses existing prompt-based ones.
CPrompt consists of two modules: Classifier Consistency Learning (CCL) and Prompt Consistency Learning (PCL).
Specifically, CCL is proposed to address the classifier inconsistency issue by exposing the current task prompt training to all classifiers seen so far.
On the other hand, PCL is proposed to handle the prompt inconsistency problem. The classifier of the current task is trained with prompts randomly selected from the pool.
It enables the classifier to better adapt to different prompts for more robust prediction.
Moreover, a multi-key mechanism is proposed to boost the prompt selection accuracy in PCL.
The comparison between existing prompt-based approaches with our Consistent Prompting is illustrated in Figure~\ref{fig:secondpic}.
We evaluate the proposed method on four challenging continual learning benchmark datasets. Extensive analysis is also provided to demonstrate the superior performance of CPrompt mainly comes from its consistent training and testing.
The contributions of this work are summarized in three-fold:
\begin{enumerate}
    \item The inconsistency issues between the training and testing of prompt-based rehearsal-free continual learning methods are identified and discussed for the first time in this work. A novel Consistent Prompting (CPrompt) is then proposed for better consistency.
    \item To maintain the classifier consistency at testing, the prompt should be exposed to all seen classifiers during training, as proposed in Classifier Consistency Learning (CCL) of CPrompt.
    \item We propose Prompt Consistency Learning (PCL) in CPrompt with two complementary purposes. For more robust testing predictions, the current classifier should be trained under different prompts. For more precise prompt selection, a multi-key mechanism is exploited.
\end{enumerate}

\section{Related Work}
\subsection{Continual Learning}
Continual learning methods aim to reduce catastrophic forgetting~\cite{mccloskey1989catastrophic,lee2017overcoming} of the old knowledge when adapting to new one. Three kinds of pipelines are proposed based on distinctive perspectives.
Regularization-based approaches~\cite{li2017learning,aljundi2018memory,kirkpatrick2017overcoming,zenke2017continual} aim to prevent significant changes to important attributes to protect previously learned knowledge from excessive interference.
As a prevalent regularization method, knowledge distillation~\cite{feng2022overcoming,wu2019large,buzzega2020dark} is widely adopted to transfer the retained knowledge in the previous model (as a teacher) to the current student model.
Parameter isolation methods~\cite{li2019learn, yan2021dynamically, ke2020continual} freeze specific parameters while allocating the rest for subsequent tasks or expanding the network for new knowledge learning.
Such approaches are inherently intuitive and can yield promising results when an ample number of parameters are extended. 
However, these techniques often induce model complexity, posing maintenance challenges.
{Rehearsal}~\cite{chen2023dynamic,bonicelli2022effectiveness,hayes2021replay,jeeveswaran2023birt,lin2023pcr} is a popular strategy in continual learning, allowing the model to partially access previous exemplars.
These approaches can effectively mitigate the forgetting of prior knowledge. However, they require additional memory and computational overhead and raise data privacy concerns as well.
{The aforementioned paradigms are highly complementary and can be combined to enhance continual learning performance~\cite{yan2021dynamically,wang2022foster,wang2022beef,chen2023dynamic}.}

\begin{figure*}[t]
  \centering
  \centerline{\includegraphics[width=0.99\textwidth]{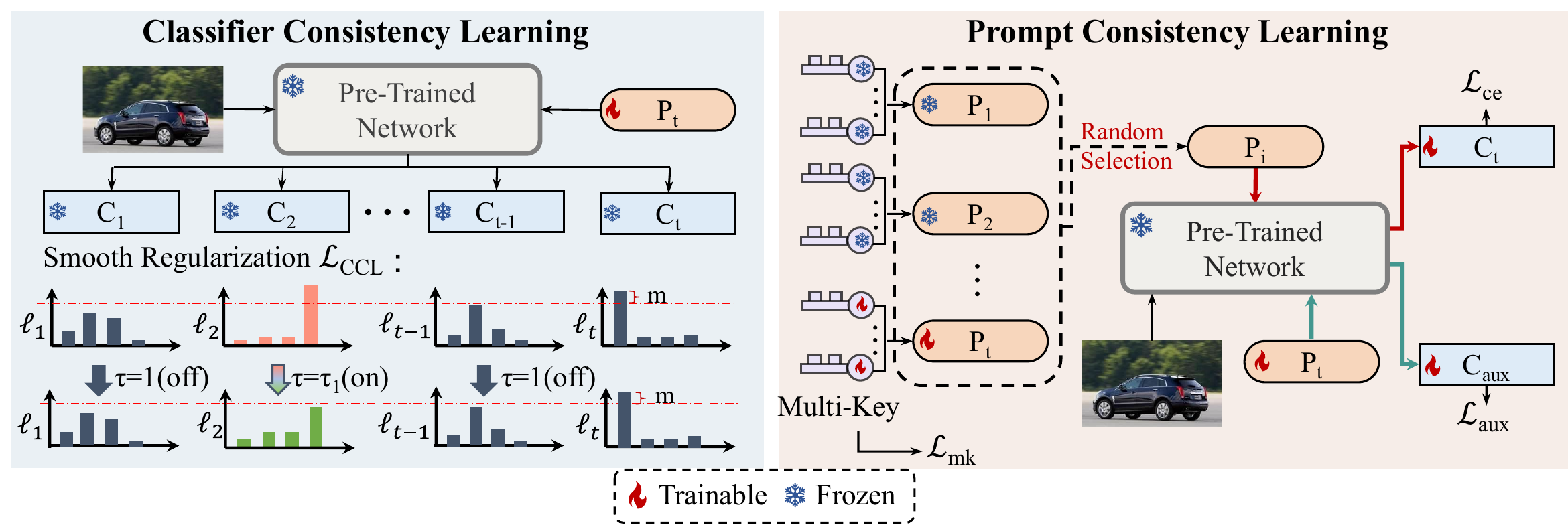}}
\caption{
{The illustration of the proposed consistent prompting (CPrompt). CPrompt aims to align the training of prompts and classifiers with testing for more consistency.
It consists of two main modules: classifier consistency learning (CCL, detailed in Section~\ref{sec:CCL}) and prompt consistency learning (PCL, detailed in Section~\ref{sec:PCL}).}
}
\label{fig:p34}
\end{figure*}

\subsection{Prompt-based Methods in Continual Learning}
Prompt~\cite{lester2021power,liu2023pre} is a fundamental technique used in Natural Language Processing (NLP). It serves as a transfer approach or provides specific instructions for downstream tasks.
{Recent prompt-based continual learning methods}~\cite{moon2023online,khan2023introducing,razdaibiedina2023progressive} handle the rehearsal-free setting by encoding {task-specific} knowledge within prompts. It enables the network to efficiently retrieve previous information by querying the appropriate prompts and eliminates the need for rehearsal buffers.
L2P~\cite{wang2022learning} proposes to learn a prompt pool and use a query-key mechanism to select a prompt. However, since the entire prompt pool is always trainable, forgetting of prior knowledge within the prompt pool is inevitable.
Instead of using the same prompt pool across tasks, two complementary pools, G-Prompt and E-Prompt, are proposed by DualPrompt~\cite{wang2022dualprompt} to encode task-invariant and tasks-specific knowledge, respectively.
CODAPrompt~\cite{smith2023coda} introduces a decomposed attention-based prompting method and expands the prompt component according to different tasks.
To handle the expansion of classifiers in continual learning, ESN~\cite{wang2023isolation} proposes a prompt-based method that employs energy self-normalization. This is achieved by adding a self-attention block, which produces consistent and high confidence scores for in-task data.

Existing prompt-based methods face difficulties in clarifying the relationship between the newly optimized prompts and the ones learned previously. This leads to interference among different prompts or components during the training process. Additionally, these prompts are not fully aligned with either input images or overall classifiers, causing the training-testing inconsistency. To address these challenges, we propose consistent prompting as a solution to enhance the effectiveness of continual learning.

\section{Consistent Prompting}
\label{sec:formatting}
\subsection{Prerequisites}

\noindent\textbf{Continual Learning Setting}\quad
Continual learning is designed to acquire knowledge from a data stream comprising T non-overlapping sequential datasets, denoted as $\mathcal{D}=\{\mathcal{D}_1,\mathcal{D}_2,...,\mathcal{D}_T\}$. Each dataset $\mathcal{D}_t$ corresponds to a specific task $t$ and can be represented as the union of individual class datasets, i.e., $\mathcal{D}_t=\bigcup_{j} \mathcal{D}_{t,j}$, where $j$ denotes the $j$th class within task $t$. The objective is to train a mapping function $f$ capable of predicting $f(x)\in Y_{t,j}$ for every input $x\in \mathcal{D}_{t,j}$, where $Y_{t,j}$ represents the $j$th class within task $t$.
{
In this research, we focus on class incremental learning (CIL) where task identities are not provided during testing.
Moreover, adopting the rehearsal-free CIL setting prohibits the use of any exemplar from prior tasks.
}

\noindent\textbf{Prompt-based {Method}}\quad
A pre-trained vision transformer (ViT)~\cite{dosovitskiy2020image} comprises an embedding layer $f_e$ and multiple self-attention layers $f_i$, $i = 1, 2, ..., N$. Each image $x\in \mathcal{D}_{t,k}$ is initially processed by the embedding layer $f_e$, yielding a sequential feature $\xi_{e} = f_e(x) \in \mathbb{R}^{L\times D}$, where $L$ refers to the number of patches, i.e., token length, and $D$ denotes the embedding dimension.
During training, we keep the parameters of the pre-trained model $f_i$ and $f_e$ frozen and exclusively update additional learnable parameters, known as the prompt $P\in \mathbb{R}^{L_P\times D}$, where $L_P$ indicates the length of the prompt.
{
Specifically, a prompt $P$ can simultaneously fit with $s$ self-attention layers by splitting $P$ into $s$ segments and each segment $p_i\in \mathbb{R}^{(L_P/s)\times D}$, $i = 1,2,...,s$.
To insert a segment $p_j$ into its corresponding self-attention layer $f_i$, we extend the output $\xi_{i-1}$ of the preceding layer $f_{i-1}$ with $p_j$, resulting the input [$p_j$;$\xi_{i-1}$] for $f_i$.}
During testing, the pre-trained model collaborates with the learned prompt.
However, task-specific prompts are learned across tasks in continual learning. 
Therefore, an additional prompt selection mechanism is required.

\subsection{Classifier Consistency Learning}\label{sec:CCL}

Prompt-based approaches typically utilize all task classifiers for inference during test time. 
However, different classifiers are learned within the corresponding task. The alignment among their behaviors is not guaranteed.
The proposed Classifier Consistency Learning (CCL) aims to handle this issue by exposing the current task prompt training to all classifiers encountered, which is straightforward and effective.
{To achieve accurate predictions, we need to ensure that each input has the highest logit response for its corresponding class among all encountered classifiers.
However, we observe that such classifiers without regularization are naturally biased towards the previous parts, resulting in higher logit values for old task classifiers than for the current one.}
{Therefore, a new regularization loss is proposed to encourage the maximum logit value of the old task classifiers to be lower than that of the current class by a predefined margin.
This regularization is applied to the current task prompt to mitigate bias and enable model training. Further details are provided below.}

{Assuming the pre-trained network as $f_\theta$, we can extract the image feature with respect to a specific class using the token [class] as in \cite{dosovitskiy2020image}}:
\begin{equation}
    h = f_\theta(x, P_t)[0],
    \label{eq:getfeature}
\end{equation}
where $x \in D_t$ is an input image of the current training task $t$, {$0$ means indexing the first [class] token.}
The extracted feature $h$ is then fed into all task classifiers, and the respective logit values are obtained via,
\begin{equation}
    \ell_i = C_i(h), i \in \{1,...,t\}.
    \label{eq:getlogit}
\end{equation}

{The proposed smooth regularization is built on the entropy of previous tasks. For the $i$th task, $i \in \{1,...,t-1\}$, 
an adaptive} entropy is calculated based on logit $\ell_i$,
\begin{equation}
\begin{split}
    \mathcal{L}_{e}(i) = - <\sigma(\ell_i / \tau),\ \log(\sigma(\ell_i))>,
\end{split}
\label{eq:2}
\end{equation}
where $\sigma$ is the softmax function and $<,\ >$ is the inner product operator. 
{We also find that blocking the gradients from $\sigma(\ell_i / \tau)$ stabilizes the optimization.}
The temperature $\tau$ is chosen accordingly,
\begin{equation}
    \tau =
    \begin{cases}
            \tau_1,\quad
            \max(\ell_i) + m \geq \max(\ell_t),
        \\
        1,\quad\text{otherwise}.
    \end{cases}
    \label{eq:temperature}
\end{equation}
where smooth regularization is needed when the maximum logit of the previous task classifier $\max(\ell_i)$ exceeds the maximum logit of the current task classifier $\max(\ell_t)$ by a margin of $m\geq0$. Therefore, $\tau$ equals to a predefined $\tau_{1}>1$ and enables the regularization.
Otherwise, smooth regularization is not necessary. Simply setting $\tau$ to $1$ yields zero gradients can turns it off. The supplementary material provides relevant theoretical analysis and proof.
An illustration of the smooth regularization process is depicted in Figure~\ref{fig:p34} for better understanding.

The Classifier Consistency Learning loss $\mathcal{L}_{CCL}$ is given by,
\begin{equation}
\mathcal{L}_{CCL} = \frac{\alpha}{t-1} \sum_{i=1}^{t-1}{ \mathcal{L}_{e}(i)},
\label{eq:3}
\end{equation}
where $\alpha$ indicates the strength of the regularization
and $t-1$ normalizes the growth of tasks in continual learning. 

\subsection{Prompt Consistency Learning}\label{sec:PCL}
{Another significant problem of prompt-based continual learning is the model's uncertainty in selecting the correct prompt for inference.}
{Therefore, the proposed Prompt Consistency Learning (PCL) aims to establish a more robust prompt-classifier relationship, ensuring correct output even with incorrect prompts.}
During training, a task-specific prompt $P_i$ is randomly selected from the current prompt pool,
\begin{equation}
    P_i, i \sim \operatorname{Uni}(1, t),
\end{equation}
where $\operatorname{Uni}(1, t)$ is a uniform distribution on the integers $1, 2, ..., t$.
All prompts except $P_t$ are frozen to preserve the encoded knowledge and defy catastrophic forgetting.
The output logit of the current task classifier $C_t$ can then be obtained via,
\begin{equation}
\ell_t = C_t(f_\theta(x, P_i)[0]).
\label{eq:4}
\end{equation}
The corresponding loss is calculated,
\begin{equation}
\mathcal{L}_{ce} = \text{CrossEntropy}(\ell_t, y),
\label{eq:5}
\end{equation}
where $y$ is the ground truth class label of the input image $x$.
{In this way, the classifier is trained to make the correct predictions even based on the wrong prompts, which is more consistent with testing.}
This training process is illustrated in Figure~\ref{fig:p34}.

It is noteworthy that the current task-specific prompt $P_t$ is selected with a probability of only $1/t$, which results in inadequate training for $P_t$. Therefore, we employ an auxiliary classifier, $C_{aux}$, to assist in the training of $P_t$ as follows:
\begin{equation}
\mathcal{L}_{aux} = \text{CrossEntropy}(C_{aux}(h), y).
\label{eq:6}
\end{equation}
Here, $h$ represents the extracted feature calculated by Eq.~(\ref{eq:getfeature}).
It is noted that we employ the auxiliary classifier on $C_{aux}$ instead of $C_t$. Employing the auxiliary classifier on $C_t$ would lead $C_t$ to lean towards adapting more to the current prompt $P_t$ rather than adapting to any task prompt with equal probability as in Eq.~(\ref{eq:5}). This could result in $C_t$ lacking robustness when selecting other prompts during testing and harming the training-testing consistency.
More discussions and results about the necessity of $C_{aux}$ is presented in Section~\ref{sec:detail_analysis}.

\noindent\textbf{{Multi-Key for Prompt Selection}}\quad
{A new multi-key mechanism is proposed for more accurate prompt selection and thus enhances the continual learning performance.}
The query feature is extracted from the pre-trained network,
which (the feature vector corresponding to [class] token~\cite{dosovitskiy2020image}) can be expressed as,
\begin{equation}
    q = f_\theta(x)[0].
\end{equation}
Query features extracted from the pre-trained network of the same class tend to be similar, while they may exhibit diversity across different classes within the same task.
Therefore, relying on a single key to represent each task, as the previous prompt-based methods did, is not sufficient.
The proposed multi-key mechanism employs multiple keys {in a prompt} to map each task, resulting in more precise representations of the various categories within each task, as illustrated in Figure~\ref{fig:p34}.
{Specifically, each class within a task is assigned a unique key, resulting in the same number of keys as classes.}
The cosine similarity is then employed to measure the discrepancy between a query and its corresponding key,
\begin{equation}
    d_{i,j}=\cos(q,k_{i,j}),
\end{equation} 
$k_{i,j}$ indicates the key of the $j$th class within task $i$.
The query retrieves the closest key and the corresponding prompt is selected,
\begin{equation}
i=\mathop{\arg\max}\limits_{i\in\{1:t\},j\in\{1:|Y_i|\}}d_{i,j},
\end{equation}
where $|Y_i|$ is the number of classes in task $i$.
During training, the softmax cross-entropy is employed to maximize the similarity between the query and its corresponding key while minimizing other similarities,
\begin{equation}
    \mathcal{L}_{mk}=-\log({\frac{e^{d_{t,y}(x)}}{\sum_{i\in\{1:t\},j\in \{1:|Y_i|\}}e^{d_{i,j}(x)}}}),
\label{eq:qk}
\end{equation}
where $y$ represents the class label of the input image $x$ at current task $t$. 
The total loss function of Prompt Consistency Learning is,
\begin{equation}
\mathcal{L}_{PCL}=\mathcal{L}_{ce}+\mathcal{L}_{aux}+\mathcal{L}_{mk}.
\end{equation}

\begin{table*}[t]
\centering
\caption{
The continual learning results on Split StanfordCars under 10-task and 20-task settings. Best results are marked in \textbf{bold}.
}
\begin{tabular}{c|ccc|ccc}
\hline
\multicolumn{1}{c|}{\multirow{2}{*}{Method}} & \multicolumn{3}{c|}{10-task} & \multicolumn{3}{c}{20-task}\\
\cline{2-7} 
  & Last-acc$\uparrow$ & Avg-acc~$\uparrow$ & FF~$\downarrow$ & Last-acc~$\uparrow$ & Avg-acc~$\uparrow$ & FF~$\downarrow$\\
\hline
UB & 83.96 & - & - & 83.96 & - & - \\
\hline
 L2P  & 60.39$\pm$1.99 & 71.92$\pm$1.12 & \textbf{13.00$\pm$0.12} 
 & 45.14$\pm$4.33 & 60.03$\pm$3.56 & \textbf{15.22$\pm$1.28} \\
 DualPrompt  & 57.27$\pm$0.34 & 70.36$\pm$2.33 & 16.31$\pm$0.97 
 & 43.99$\pm$1.55 & 60.22$\pm$0.86 & 18.25$\pm$1.45\\
 ESN  & 56.91$\pm$0.56 & 72.82$\pm$0.79 & 13.50$\pm$1.64 
& 46.53$\pm$2.02 & 62.54$\pm$1.81 &15.96$\pm$0.76  \\
CODAprompt & 62.24$\pm$0.14 & 73.28$\pm$0.93 & 15.08$\pm$0.89 
 &48.94$\pm$1.77 &63.78$\pm$1.46 & 17.38$\pm$2.16 \\
Ours  & \textbf{66.77$\pm$0.37} & \textbf{76.81$\pm$0.27} & 13.95$\pm$0.46 
 & \textbf{55.16$\pm$0.19} & \textbf{68.74$\pm$0.83} & 20.02$\pm$1.78\\
\hline
\end{tabular}
\label{cars196}
\end{table*}

\begin{table*}[t]
\centering
\caption{
The continual learning results on Split ImageNet-R under 10-task and 20-task settings. Best results are marked in \textbf{bold}.
}
\begin{tabular}{c|ccc|ccc}
\hline
\multicolumn{1}{c|}{\multirow{2}{*}{Method}} & \multicolumn{3}{c|}{10-task} & \multicolumn{3}{c}{20-task}\\
\cline{2-7} 
  & Last-acc~$\uparrow$ & Avg-acc~$\uparrow$ & FF~$\downarrow$  & Last-acc~$\uparrow$ & Avg-acc~$\uparrow$ & FF~$\downarrow$\\
\hline 
UB  & 80.27 & - & -  & 80.27 & - & - \\
\hline
L2P  & 74.60$\pm$0.90 & 80.83$\pm$1.39 & \textbf{4.52$\pm$0.50} 
 & 72.09$\pm$1.12 & 78.39$\pm$0.94 & \textbf{4.86$\pm$1.37} \\
 DualPrompt  & 74.87$\pm$0.85 & 81.39$\pm$1.25 & 7.18$\pm$0.15 
 & 71.69$\pm$1.06 & 79.12$\pm$1.27 & 7.68$\pm$0.96 \\
 ESN  & 75.11$\pm$0.36 & 81.63$\pm$1.10 & 5.68$\pm$0.77 
 & 70.57$\pm$0.62 & 77.95$\pm$0.76 & 6.84$\pm$0.36 \\
CODAprompt  & 75.51$\pm$0.81 & 81.32$\pm$1.01 & 5.91$\pm$1.36 
 & 72.25$\pm$0.78 & 78.07$\pm$0.40 & 6.65$\pm$0.31\\

Ours  & \textbf{77.14$\pm$0.11} & \textbf{82.92$\pm$0.70} & 5.97$\pm$0.68
 & \textbf{74.79$\pm$0.28} & \textbf{81.46$\pm$0.93} & 7.34$\pm$0.65\\
\hline
\end{tabular}
\label{imagenetr}
\end{table*}

Finally, the overall learning objective of the proposed CPrompt is,
\begin{equation}
    \mathcal{L}=\mathcal{L}_{CCL} + \mathcal{L}_{PCL}.
\end{equation}

\section{Experiments}

\subsection{Experimental Details}

\noindent\textbf{Datasets and Protocols}\quad
Extensive experiments on four benchmark datasets are conducted for thorough comparisons among different continual learning methods. Following the class incremental setting, all test samples are without task identity.
\begin{itemize}
\item \textbf{StanfordCars}~\cite{krause20133d} is a fine-grained car dataset comprising 196 classes and 16,185 images. There are 8,144 training images and the rest for testing. 
All classes are randomly divided into the 10-task (20 classes per task) and 20-task (10 classes per task) continual learning setting, denoted Split StanfordCars.
It poses a challenge due to the distinct image styles and the difficulty in distinguishing between fine-grained classes.
\item  \textbf{ImageNet-R}~\cite{hendrycks2021many} has 30,000 images of 200 ImageNet classes.
The images of each class exhibit various styles, including art, cartoons, Deviant-Art, graffiti, and hard examples sourced from the original ImageNet dataset. This benchmark is challenging because the various styles significantly differ from the pre-training data.
The continual learning benchmark, Split ImageNet-R, consists of the 10-task (20 classes per task) and 20-task (10 classes per task) settings.

\item \textbf{DomainNet}~\cite{peng2019moment} dataset is a cross-domain dataset, including 345 common objects from 6 diverse domains, including Clipart, Real, Sketch, Infograph, Painting, and Quickdraw. Due to the significant disparity in image counts of different classes, the 200 categories with the most images are selected.
Its continual learning setting consists of 10 tasks (10 classes per task). Different classes are distributed into different tasks at random. 
Moreover, each task comprises images from multiple domains rather than a single one as in the existing split~\cite{wang2023isolation}.

\item \textbf{CIFAR-100}~\cite{krizhevsky2009learning} consists of 60,000 32 × 32 colour images of 100 classes. There are 500 training images and 100 test images per class.
It has a widely adopted continual learning benchmark with 10 tasks (10 classes per task).

\end{itemize}

\noindent\textbf{Evaluation Metrics}\quad
The continual learning performance of classification models is mainly evaluated by two metrics~\cite{wang2023comprehensive}: the average accuracy of all classes after learning the last task (denoted as Last-acc) and the averaged incremental accuracy over all learned tasks (denoted as Avg-acc).
The average forgetting~\cite{wang2023comprehensive}, denoted as FF, provides additional context about the performance drops over tasks.
We give more emphasis to Last-acc and Avg-acc as they are more comprehensive metrics~\cite{smith2023coda}.

\begin{figure*}[t]
  \centering
  \centerline{\includegraphics[width=0.99\linewidth]{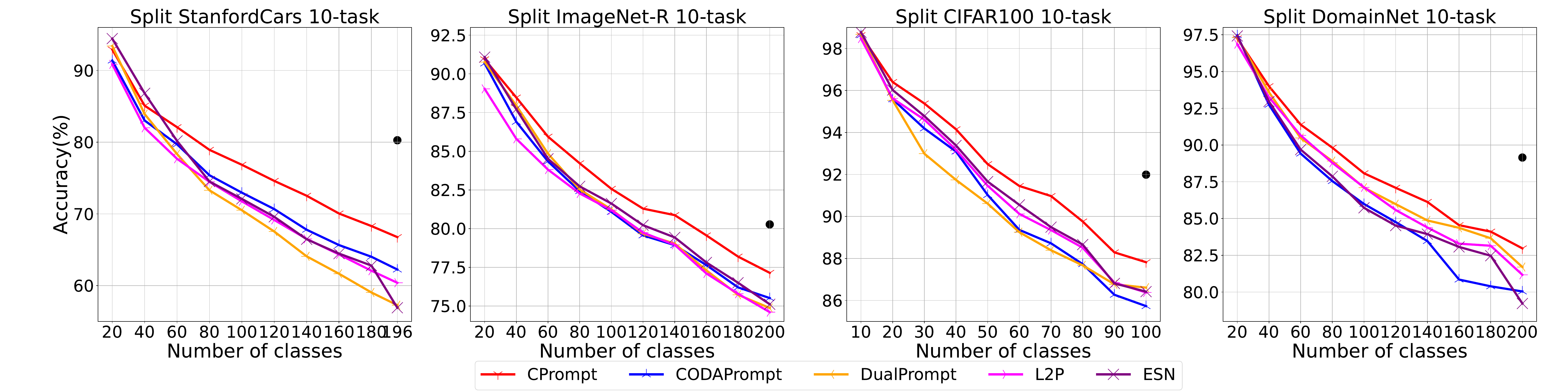}}
\caption{
Illustrations of continual learning performance at each task. Each dot indicates the accuracy of the seen classes.
{The result of the upper-bound (UB) is represented by a dot on the overall classes.}
}
\label{fig:curve}
\end{figure*}

\noindent\textbf{Implementation Details}\quad
Our model training employs the SGD optimizer with a momentum of 0.9 and an initial learning rate of 0.01. The initial learning rate gradually diminishes to zero following a cosine annealing scheduler. The mini-batch size is 16.
{In practice, the balancing hyper-parameters of different losses are all fixed at 1 without tuning.}
{The hyper-parameters $(\tau_1, m)$ in Eq.~(\ref{eq:temperature}) are $(1.02, 0.05)$ for DomainNet, $(1.2, 0)$ for CIFAR-100 and $(1.15, 0.1)$ otherwise.} 
These values are chosen via cross-validation.
{Following the previous setting~\cite{wang2022learning,wang2022dualprompt}}, the ImageNet pre-trained ViT-B/16~\cite{dosovitskiy2020image} is used as the backbone. Moreover, the ViT-B/16 architecture encompasses 12 self-attention layers. We divided each task-specific prompt into two segments and inserted them into the first and the middle 6th self-attention layers, respectively.
All results are averaged over 3 runs with the corresponding standard deviations reported to mitigate the influence of random factors.

\noindent\textbf{Competitors}\quad
We focus on comparing our model with the state-of-the-art prompt-based methods, i.e., L2P~\cite{wang2022learning}, DualPrompt~\cite{wang2022dualprompt}, ESN~\cite{wang2023isolation} and CODAprompt~\cite{smith2023coda}. 
For fair comparisons, their best results are also reproduced under our experimental settings.
The upper-bound (UB) performance is achieved by fine-tuning the prompt and classifier with all task data collectively.

\begin{table}[t]
\centering
\caption{
The continual learning results on Split CIFAR-100 under the 10-task setting. Best results are marked in \textbf{bold}.
}
\begin{tabular}{c|ccc}
\hline
Method & Last-acc~$\uparrow$ & Avg-acc~$\uparrow$ & FF~$\downarrow$ \\
\hline
UB & 91.99 & - & -  \\
\hline
L2P  & 86.38$\pm$0.31 & 91.45$\pm$0.19 & 5.88$\pm$0.76 \\
DualPrompt  & 86.61$\pm$0.22 & 90.82$\pm$1.47 & 5.86$\pm$0.62 \\
ESN  & 86.42$\pm$0.80 & 91.65$\pm$0.67 & 6.08$\pm$0.48 \\
CODAprompt & 85.73$\pm$0.14 & 91.03$\pm$0.57 & 7.13$\pm$0.44 \\
Ours & \textbf{87.82$\pm$0.21} & \textbf{92.53$\pm$0.23} & \textbf{5.06$\pm$0.50} \\
\hline
\end{tabular}
\label{cifar100}
\end{table}

\subsection{Main Results}
Comparisons between our Consistent Prompting (CPrompt) with its SOTA competitors on the Split StanfordCars benchmarks are shown in Table~\ref{cars196}.
The Split StanfordCars benchmark is a challenging task for continual learning methods, as their results are far behind the upper bound of joint learning and non-neglectable forgetting occurs.
However, the proposed CPrompt achieves the best performance among its counterparts.
Specifically, CPrompt outperforms the SOTA CODAprompt significantly, achieving $4.53\%$ and $3.53\%$ higher Last-acc and Avg-acc respectively under the 10-task setting.
Such improvements become more apparent with additional tasks.
Under the 20-task setting, CPrompt achieves $6.22\%$ higher Last-acc and $4.96\%$ higher Avg-acc than those of CODAprompt.

Our CPrompt and SOTA methods are also compared in Split ImageNet-R, as shown in Table~\ref{imagenetr}.
Under the 10-task setting, our method still outperforms CODAprompt, showing about $1.60\%$ improvements in both Last-acc and Avg-acc.
Larger improvements, $2.54\%$ higher Last-acc and $3.33\%$ higher Avg-acc, achieved by CPrompt are also observed under the more challenging 20-task setting, similar to the trends above.

Table~\ref{cifar100} and Table~\ref{domainnet} present the results of Split CIFAR-100 and Split DomainNet both under 10-task continual learning.
The proposed CPrompt consistently outperforms other prompt-based methods on both benchmarks under all criteria.
Furthermore, CPrompt's improvements are non-trivial, mostly resulting in a greater than $1\%$ increase in Last-acc and Avg-acc compared to all competitors.

\begin{table}[t]
\centering
\caption{
The continual learning results on Split DomainNet under the 10-task setting. Best results are marked in \textbf{bold}.
}
\begin{tabular}{c|ccc}
\hline
Method  & Last-acc~$\uparrow$ & Avg-acc~$\uparrow$ & FF~$\downarrow$ \\
\hline
UB  & 89.15 & - & - \\
\hline
L2P  & 81.17$\pm$0.83 & 87.43$\pm$0.95 & 8.98$\pm$1.25 \\
DualPrompt  & 81.70$\pm$0.78 & 87.80$\pm$0.99 & 8.04$\pm$0.31 \\
ESN  & 79.22$\pm$2.04 & 86.69$\pm$1.18 & 10.62$\pm$2.12 \\
CODAprompt & 80.04$\pm$0.79 & 86.27$\pm$0.82 & 10.16$\pm$0.35 \\
Ours  & \textbf{82.97$\pm$0.34} & \textbf{88.54$\pm$0.41} & \textbf{7.45$\pm$0.93} \\
\hline
\end{tabular}
\label{domainnet}
\end{table}

\begin{table*}[t]
\centering
\small
\caption{
Ablation study of the proposed CPrompt with three components: Classifier Consistency Learning (CCL), Prompt Consistency Learning without Multi-Key mechanism
({PCL w/o MK})
, and Multi-Key (MK) mechanism.
}
\begin{tabular}{ccc|cc|cc|cc|cc}
\hline
\multicolumn{1}{c}{\multirow{3}{*}{CCL}} 
&\multicolumn{1}{c}{\multirow{3}{*}{\begin{tabular}[c]{@{}c@{}}{PCL}\\{w/o MK}\end{tabular}}} 
&\multicolumn{1}{c|}{\multirow{3}{*}{MK}} & \multicolumn{4}{c|}{Split StanfordCars} & \multicolumn{4}{c}{Split ImageNet-R}\\
\cline{4-11}
\multicolumn{3}{c|}{} & \multicolumn{2}{c|}{10-task} & \multicolumn{2}{c|}{20-task}
& \multicolumn{2}{c|}{10-task} & \multicolumn{2}{c}{20-task}\\
\cline{4-11}
& & & Last-acc~$\uparrow$ & Avg-acc~$\uparrow$  & Last-acc~$\uparrow$ & Avg-acc~$\uparrow$ & Last-acc~$\uparrow$ & Avg-acc~$\uparrow$  & Last-acc~$\uparrow$ & Avg-acc~$\uparrow$\\

\hline
& &  & 61.96 & 73.10
& 49.23 & 64.10 
& 74.75 & 81.19
& 72.44 & 79.80  \\
\ding{52} & & & 62.52 & 73.42 
& 51.89 & 66.11 
 & 76.37 & 82.19 
& 73.63 & 80.46 \\
 &\ding{52} & & 65.18 & 75.82
& 52.55 & 67.96
& 75.93 & 81.93
& 73.83 & 81.02  \\
& &\ding{52} & 63.66 & 74.26
& 51.65 & 65.91
& 75.36 & 82.01
& 73.14 & 80.22  \\
\ding{52} &\ding{52} & & 66.37 & 76.58 
 & 53.44 & 67.16
 & 76.53 & 82.61 
& 74.51 & 81.43  \\
\ding{52} &\ding{52} &\ding{52} & 66.77 & 76.81 
& 55.16 & 68.74 
& 77.14 & 82.92 
& 74.79 & 81.46  \\
\hline
\end{tabular}
\label{ab_cars196}
\end{table*}

\begin{table}[t]
\centering
\caption{
Detail analysis of CCL on 10-task continual learning of Split StanfordCars.
{Reported results are obtained by matching each input with the corresponding prompt.}
w/o CCL means CPrompt without CCL.
}
\begin{tabular}{c|ll}
\hline
&Last-acc~$\uparrow$ &Avg-acc~$\uparrow$\\
\hline
w/o CCL & 67.78 &77.27 \\
CPrompt & 71.14\textcolor{red}{\fontsize{8}{9}\selectfont +3.36} &80.44\textcolor{red}{\fontsize{8}{9}\selectfont +3.17} \\
\hline
\end{tabular}
\label{da_classifier_consistency}
\end{table}

Detailed comparisons between different methods along the continual learning procedure are illustrated in Figure~\ref{fig:curve}.
It demonstrates the superiority of our method.
The curves of CPrompt are consistently above those of its counterparts across different tasks.
Typically, these performance gaps tend to widen as tasks become more complex and involve greater numbers of classes.
This finding suggests that CPrompt can be more resistant to catastrophic forgetting compared to its SOTA competitors.

\subsection{Ablation Study}
The proposed CPrompt consists of two main components: Classifier Consistency Learning (CCL) and Prompt Consistency Learning (PCL) including the Multi-Key (MK) mechanism.
The effectiveness of each component is evaluated through ablative experiments
on two large-scale continual learning benchmarks, Split StanfordCars and Split ImageNet-R.
The results are presented in Table~\ref{ab_cars196}.
The performance of our full CPrompt (with all three components) is significantly better than that of the vanilla backbone (results in the first row). For example, CPrompt achieves $4.81\%$ higher Last-acc and $3.71\%$ higher Avg-acc than the backbone under the Split StanfordCars 10-task setting.
Furthermore, including each component alone can typically result in substantial improvements, exceeding $1\%$, in both criteria. Therefore, the effectiveness of each proposed component is demonstrated.
Last but not least, combining CCL and PCL {w/o MK} improves the performance of each individual component. Such a combination clearly boosts the performance of all settings to the second best. It demonstrates that {these two components} serve distinctive purposes and complement each other.

\begin{table}[t]
\centering
\caption{Results of DualPrompt with CCL on 10-task continual learning of Split StanfordCars.}
\begin{tabular}{c|ll}
\hline
& Last-acc~$\uparrow$ &Avg-acc~$\uparrow$ \\
\hline
DualPrompt   & 57.27 & 70.36 \\
+CCL  & 61.79\textcolor{red}{\fontsize{8}{9}\selectfont+4.52} 
& 72.07\textcolor{red}{\fontsize{8}{9}\selectfont+1.71} \\
\hline
\end{tabular}
\label{da_ccl_baseline}
\end{table}

\subsection{Detailed Analysis}\label{sec:detail_analysis}

\noindent\textbf{Detailed Analysis of CCL}\quad
During the training phase, CCL is proposed to regulate the behavior of all classifiers, aiming to achieve more consistency across training and testing. In order to demonstrate the advantages exclusively derived from the consistent classifier, the corresponding prompt of each input is provided during testing. Table~\ref{da_classifier_consistency} presents the corresponding comparison between the CPrompts trained with and without CCL. 
Our full approach (with CCL) achieves significantly better results, {$3.36\%$ higher Last-acc and $3.17\%$ higher Avg-acc,}
than the CPrompt without CCL.
Furthermore, CCL can be applied to other prompt-based continual learning methods and brings them significant improvements, e.g., $4.52\%$ Last-acc, to DualPrompt, as demonstrated in Table~\ref{da_ccl_baseline}. More results of CODAPrompt and L2P with our approach are in the supplementary material.

\noindent\textbf{Detailed Analysis of PCL}\quad
{To demonstrate the benefits of robust prediction brought by PCL, a new experiment procedure is developed. During testing, we consistently choose the prompt of the initial task, which does not correspond to any of the subsequent tasks. Using this prompt, we calculate the model's performance from the second task to the last one and then take the average. This setting helps to showcase the model's robustness in dealing with mismatched prompts, where existing methods may have difficulties.}
Table~\ref{da_pcl} presents the corresponding comparison between the CPrompts trained with and without PCL. Our full CPrompt trained with PCL outperforms its counterpart with large margins, $10.92\%$ and $7.54\%$ on Last-acc and Avg-acc, respectively. 
This clearly demonstrates the necessity of PCL in the proposed CPrompt.
Moreover, combining PCL with the DualPrompt method yields significant improvements, as demonstrated in Table~\ref{dualprompt_pcl}.

\begin{table}[t]
\centering
\caption{
Detail analysis of PCL on 10-task continual learning of Split StanfordCars.
{With the prompt selection fixed at the initial task one, the reported results are averaged from the second to the final task.}
w/o PCL means CPrompt without PCL.
}
\begin{tabular}{c|ll}
\hline
& Last-acc~$\uparrow$ &Avg-acc~$\uparrow$ \\
\hline
w/o PCL & 48.83 &63.17 \\
CPrompt & 59.75\textcolor{red}{\fontsize{8}{9}\selectfont+10.92}
&70.71 \textcolor{red}{\fontsize{8}{9}\selectfont+7.54}
\\
\hline
\end{tabular}
\label{da_pcl}
\end{table}

\begin{table}[t]
\centering
\caption{Results of DualPrompt with PCL on 10-task continual learning of Split StanfordCars.}
\begin{tabular}{c|ll}
\hline 
& Last-acc~$\uparrow$ &Avg-acc~$\uparrow$ \\
\hline
DualPrompt & 57.27 &70.36 \\
+PCL & 60.09\textcolor{red}{\fontsize{8}{9}\selectfont+2.82}
&72.12 \textcolor{red}{\fontsize{8}{9}\selectfont+1.76}\\
\hline
\end{tabular}
\label{dualprompt_pcl}
\end{table}

\noindent\textbf{Detailed Analysis of MK}\quad
The proposed multi-key (MK) mechanism assigns multiple keys to each task prompt to handle the inherent diversity of different classes.
This approach leads to higher accuracy, i.e., for more than $20\%$, in selecting task-specific prompts, thus improving continual learning performance compared to the vanilla one-key (OK) mapping per task prompt, as illustrated in Table~\ref{da_mk}.
{The importance of accurate prompt selection can also be demonstrated by the upper-bound performance under the setting of task incremental learning (TIL).
Appropriate input prompts can always be selected ($100\%$ Task-acc) with the task identities provided in TIL. As a result, its performance is clearly better than the CIL counterparts.
}

\begin{table}[t]
\centering
\caption{
Detail analysis of the Multi-Key (MK) mechanism on 10-task continual learning of Split StanfordCars.
The experiment is conducted on the vanilla backbone network with the one-key (OK) mechanism by default.
Task-acc refers to the accuracy of selecting an appropriate prompt. 
}
\begin{tabular}{c|lll}
\hline

& Task-acc~$\uparrow$ &Last-acc~$\uparrow$ &Avg-acc~$\uparrow$ \\

\hline
TIL &100 &67.50 &77.22 \\
\hline
OK   & 19.44 & 61.96 &73.10 \\
MK   & 39.83\textcolor{red}{\fontsize{8}{9}\selectfont+20.39} 
& 63.66\textcolor{red}{\fontsize{8}{9}\selectfont+1.70} 
&  74.26\textcolor{red}{\fontsize{8}{9}\selectfont+1.16}  \\
\hline
\end{tabular}
\label{da_mk}
\end{table}

\noindent\textbf{Detailed Analysis of auxiliary classifier $C_{aux}$}\quad 
The impacts of auxiliary classifier $C_{aux}$ and its corresponding loss $\mathcal{L}_{aux}$ in Eq.~(\ref{eq:6}) are evaluated in Table~\ref{da_aux_class}.
w/o $\mathcal{L}_{aux}$ refers to training CPrompt without auxiliary classifier.
w/o $C_{aux}$ indicates learning $C_t$ instead of $C_{aux}$ with the auxiliary loss $\mathcal{L}_{aux}$.
{Both variants are inferior to CPrompt, demonstrating the effectiveness of the auxiliary classifier $C_{aux}$ and its objective $\mathcal{L}_{aux}$, as shown in Table~\ref{da_aux_class}.}

\begin{table}[t]
\centering
\caption{Detail analysis of auxiliary classifier $C_{aux}$ on 10-task continual learning of Split StanfordCars.
}
\begin{tabular}{c|cc}
\hline 
& Last-acc~$\uparrow$ &Avg-acc~$\uparrow$ \\
\hline
w/o $\mathcal{L}_{aux}$ & {60.34} & {73.17} \\
w/o $C_{aux}$ & 64.17 &74.32 \\
CPrompt & 66.77 &76.81 \\
\hline
\end{tabular}
\label{da_aux_class}
\end{table}

\section{Conclusion}
In this paper, the training-testing inconsistency of existing prompt-based continual learning methods is first revealed and thoroughly discussed. We propose a novel approach, consistent prompting (CPrompt), to handle this important issue.
Our CPrompt consists of two complementary components: classifier consistency learning (CCL) and prompt consistency learning (PCL).
CCL addresses the classifier inconsistency by training with all classifiers, while PCL enhances the prediction robustness and prompt selection accuracy to alleviate the prompt inconsistency.
The effectiveness of the proposed method is demonstrated by its state-of-the-art performance. 
Extensive analysis is also conducted to show the importance of training-testing consistency in prompt-based methods.
\textbf{Limitations.} The multi-key mechanism is used to improve prompt selection accuracy during testing. It improves prompt selection accuracy by over $20\%$, as shown in Table~\ref{da_mk}. However, it still has a relatively low accuracy, i.e., less than $40\%$, achieved.
Further improving the prompt selection accuracy can significantly boost the continual learning performance, as suggested by TIL's superior performance in Table~\ref{da_mk}.
We will further enhance the proposed consistent prompting in this direction.

\noindent\textbf{Acknowledgement}\quad This research is supported by the National Natural Science Foundation for Young Scientists of China (No. 62106289).

{
    \small
    \bibliographystyle{ieeenat_fullname}
    \bibliography{main}
}

\end{document}


\maketitlesupplementary

\section{Smooth Regularization analysis}

The impact of $\tau$ on the adaptative entropy loss (Eq.~(3)) and smooth regularization process are discussed in detail.
{Suppose that there are $n_i$ classes seen within task $i$. $g_j = \sigma_j{(\ell_{i}/\tau)}$ and $\rho_j=\sigma_j{(\ell_{i})}$ refer to the probabilities of the $j$th class based on different logits. Therefore, Eq.~(3) can be rewrite as,}
\begin{equation}
\begin{aligned}
        \mathcal{L}_e(i)=&- <\sigma(\ell_i / \tau),\ \log(\sigma(\ell_i))>\\
        &=-\sum_{j=1}^{n_i}g_j\log{\rho_j}. 
\end{aligned}
\label{eg:sup}
\end{equation}
{The gradient of $\mathcal{L}_e(i)$ with respect to $\rho_j$ can be calculated,}
\begin{equation}
\begin{aligned}
    \frac{\partial\mathcal{L}_e(i)}{\partial\rho_j}&=\frac{\partial}{\partial\rho_j}(-\sum_{j=1}^{n_i}g_j\log{\rho_j})\\
    &=\frac{\partial}{\partial\rho_j}(-\sum_{j=1}^{n_i-1}g_j\log{\rho_j}-g_{n_i}\log{\rho_{n_i}})\\
    &=\frac{\partial}{\partial\rho_j}(-\sum_{j=1}^{n_i-1}g_j\log{\rho_j}-g_{n_i}\log{(1-\sum_{j=1}^{n_i-1}\rho_j)})\\
    &=-\frac{g_j}{\rho_j}+\frac{g_{n_i}}{1-\sum_{j=1}^{n_i-1}\rho_j}\\
    &=-\frac{g_j}{\rho_j}+\frac{g_{n_i}}{\rho_{n_i}},\\
\end{aligned}
\label{eg:18}
\end{equation}
{where $\rho_n=1-\sum_{j=1}^{n_i-1}\rho_j$.}
{The impact of this gradient on $\rho_j$ is,}
\begin{equation}
\begin{aligned}
    \rho_j := \rho_j - \frac{\partial\mathcal{L}_e(i)}{\partial\rho_j}.
\end{aligned}
\label{eg:grad_impact}
\end{equation}
{Given that the gradient from $g_j$, i.e., $\sigma{(\ell_j/\tau)}$, are blocked and does not flow back, $g_j=\rho_j$ and $g_{n_i}=\rho_{n_i}$ when $\tau = 1$ and the corresponding gradient $\frac{\partial\mathcal{L}_e(i)}{\partial\rho_j}$ are all zeros.
Therefore, the smooth regularization will be turned off.}
{More importantly, the proposed loss of Eq.~(3) has the desired smoothing effect by setting $\tau>1$.
Specifically, when $\rho_j$ reaches a large value, e.g. close to $1$, $\frac{g_j}{\rho_j}$ will be smaller than $\frac{g_{n_i}}{\rho_{n_i}}$, resulting in positive gradients to lower $\rho_j$ down.
In contrast, if $\rho_j$ reaches a small value, for example close to $0$, $\frac{g_j}{\rho_j}$ will be greater than $\frac{g_{n_i}}{\rho_{n_i}}$, resulting in negative gradients to lift $\rho_j$.}

\noindent\textbf{Intuitive alternatives to Smooth Regularizations}\quad
{Two smoothing techniques are proposed and compared with ours.
The first technique provides an evenly distributed label (EDL), $\frac{1}{n_i}$ on each entry, as the supervision. Its learning objective becomes,}
\begin{equation}
        \mathcal{L}_e(i)=- <\frac{1}{n_i},\ \log(\sigma(\ell_i))>.
\end{equation}
{Another smoothing baseline involves the one-hot label $\delta$ by randomly activating a category entry. This smoothing technique is denoted random one-hot, ROH, with cross-entropy objective,}
\begin{equation}
        \mathcal{L}_e(i)=- <\delta,\ \log(\sigma(\ell_i))>.
\end{equation}

Comparisons between the proposed smooth regularization and the alternative ones are shown in Table~\ref{da_sr}. Our approach outperforms the two baseline methods by a significant margin.

\begin{table}[t]
\centering
\caption{Results of different regularization methods on 10-task continual learning of Split StanfordCars.
}
\begin{tabular}{c|cc}
\hline 
& Last-acc~$\uparrow$ &Avg-acc~$\uparrow$ \\
\hline
EDL  & 39.73 & 50.93 \\
ROH & 40.99 &51.59 \\
Ours & 66.77 &76.81 \\
\hline
\end{tabular}
\label{da_sr}
\end{table}

\section{Recent relevant methods}
SLCA~\cite{zhang2023slca} and HiDe-Prompt~\cite{wang2024hierarchical} handle the classifier inconsistency with a post-hoc alignment technique to boost performance. Both methods involve storing the mean and covariance of each class feature and constructing Gaussian distributions for each class. Subsequently, unified classifiers are retrained by sampling features generated from these Gaussian distributions. However, saving the covariance of each class feature incurs significant memory overhead, particularly considering the sensitivity of continual learning to memory usage. Notably, our proposed classifier consistency learning approach only introduces regularization without requiring additional memory overhead.

\section{Analysis of the FF score}
\begin{figure}[t]
  \centering
  \centerline{\includegraphics[width=1.0\linewidth]{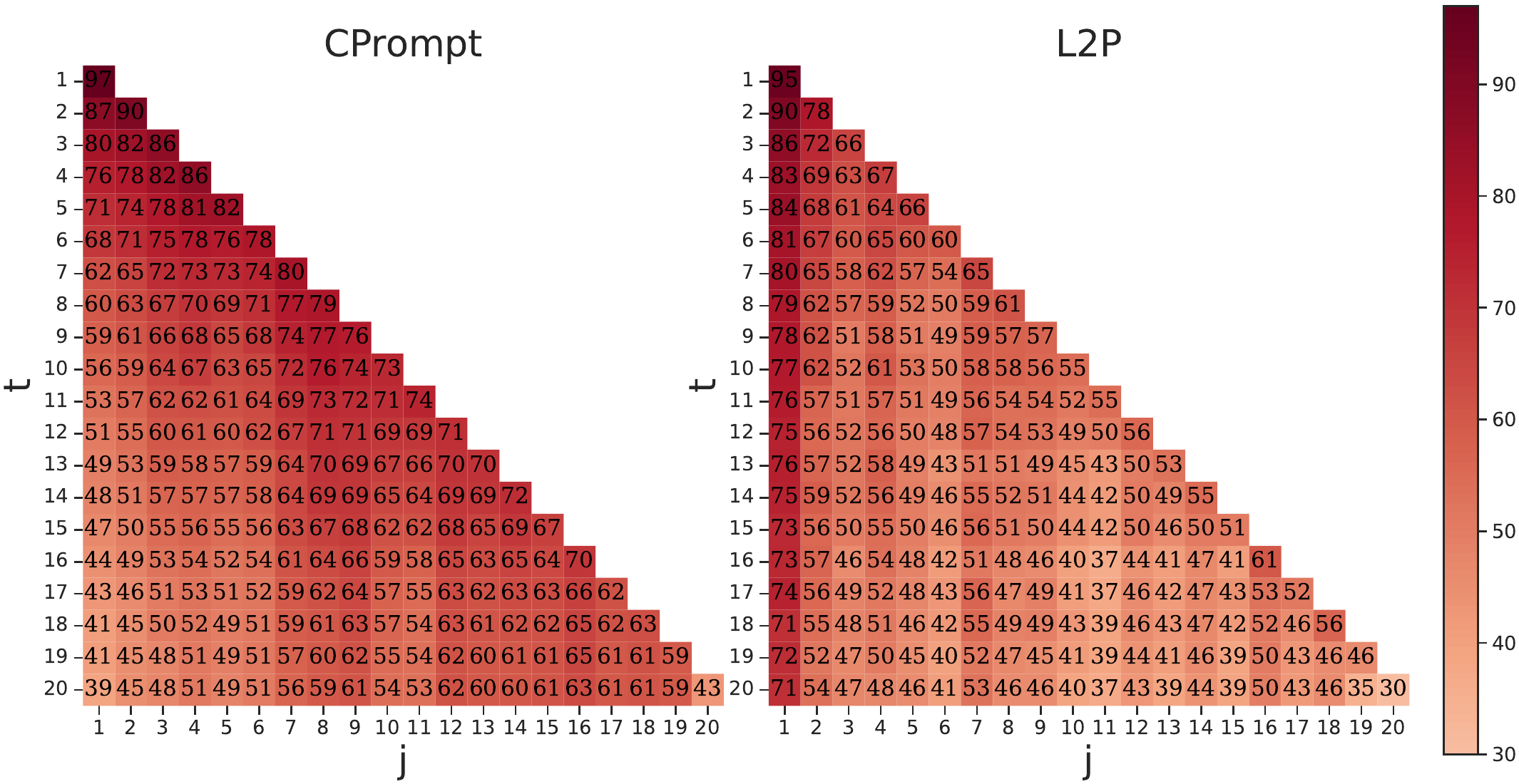}}
\caption{
Detailed illustration of FF score. All $a_{t,j}$ values are recorded and presented with a heat map. 
The deeper the red, the higher the value.
Best viewed in colors and zoom-in for the values.
}
\label{fig:heatmap}
\end{figure}

According to the FF score computation 

\begin{equation}
    FF_t = \frac{1}{t-1}\sum_{j=1}^{t-1}\{\max_{i\in\{1,...,t-1\}}(a_{i,j})-a_{t,j}\},\forall j < t,
    \label{eq:ffscore}
\end{equation}
all $a_{t,j}$ over the 20 tasks in Table 1 can be depicted as in Figure~\ref{fig:heatmap}
, where CPrompt and L2P (with the best FF score) are compared.
The $\max_{i\in\{1,...,t-1\}}(a_{i,j})$ of CPrompt are larger than those of L2P, as the diagonal area of CPrompt shows deeper red than that of L2P, resulting in larger differences and thus higher FF of CPrompt.
Moreover,
CPrompt achieves superior overall performance and stability-plasticity trade-off, as Figure~\ref{fig:heatmap} suggests.

\section{More results with our approach}

As shown in Table~\ref{table:performance}, PCL and CCL consistently improve various backbones with clear margins.

\begin{table}[h]
\centering
\footnotesize
\caption{
Continual learning results of the Split StanfordCars 10-task. 
\textcolor{black}{The improvements over the backbone are shown in red.} L: Last-acc, A: Avg-acc.
}
\label{table:performance}

\begin{tabular}{l|cc|cc}
\hline
\multicolumn{1}{c|}{\multirow{2}{*}{Method}}  & \multicolumn{2}{c|}{CODAPrompt} & \multicolumn{2}{c}{L2P} \\
\cline{2-5}
 & L & A & L & A \\
\hline
Backbone & 62.24 & 73.28 & 60.39 & 71.92  \\
\hline
+CCL & 65.12\textcolor{red}{\scriptsize{+2.88}} & 75.37\textcolor{red}{\scriptsize{+2.09}} & 63.69\textcolor{red}{\scriptsize{+3.30}} & 72.53\textcolor{red}{\scriptsize{+0.61}}   \\
+PCL & 63.49\textcolor{red}{\scriptsize{+1.25}} & 74.21\textcolor{red}{\scriptsize{+0.93}} & 61.82\textcolor{red}{\scriptsize{+1.43}} & 72.06\textcolor{red}{\scriptsize{+0.14}}  \\

+BOTH & 65.31\textcolor{red}{\scriptsize{+3.07}} & 75.81\textcolor{red}{\scriptsize{+2.53}} & 63.71\textcolor{red}{\scriptsize{+3.32}} & 72.71\textcolor{red}{\scriptsize{+0.79}}   \\
\hline

\end{tabular}
\end{table}

\section{Resource consumption}
Memory usage of a method is evaluated by the number of trainable parameters. Its computational requirement is evaluated by FLOPs.
As shown in Table~\ref{tab:param}, the proposed method consumes both resources at relatively low levels.

\begin{table}[h]
  \centering
  \footnotesize
  \caption{
  Different methods' computation and memory consumption on Split StanfordCars 10-task.
  }
  \label{tab:param}
  \begin{tabular}{c|c|c|c|c|c}
    \hline
    Method &Ours & ESN & L2P & Dual & CODA  \\
    \hline
    Params ($\times 10^6$) & 0.92 & 30.49 & 0.78 & 0.44 & 89.79  \\
    \hline
    FLOPs ($\times 10^9$) & 23.62 & 23.97 & 23.62 & 23.62 & 33.77  \\
    \hline
\end{tabular}
\end{table}

\vspace{2\baselineskip}
{
    \small
    \bibliographystyle{ieeenat_fullname}
    \bibliography{main}
}